\documentclass{article}

\newif\ifanonymous
\anonymousfalse

\ifanonymous
  \usepackage{neurips_2026}
\else
  \usepackage[preprint]{neurips_2026}
\fi

\usepackage[utf8]{inputenc}
\usepackage[T1]{fontenc}
\usepackage{hyperref}
\usepackage{url}
\usepackage{booktabs}
\usepackage{amsfonts}
\usepackage{amsmath}
\usepackage{nicefrac}
\usepackage{microtype}
\usepackage{graphicx}
\usepackage{multirow}
\usepackage{xcolor}
\usepackage{colortbl}

\title{MedDialBench: Benchmarking LLM Diagnostic Robustness \\ under Parametric Adversarial Patient Behaviors}

\ifanonymous
\author{Anonymous}
\else
\author{
  Xiaotian Luo \quad Xun Jiang\thanks{Corresponding author.} \quad Jiangcheng Wu \\
  Shanda Group \\
  \texttt{\{xiaotian.luo, jiangcheng.wu\}@thetahealth.ai} \\
  \texttt{jiangxun@shanda.com}
}
\fi

\begin{document}

\maketitle

% ============================================================
% ABSTRACT
% ============================================================
\begin{abstract}
Interactive medical dialogue benchmarks have shown that LLM diagnostic accuracy degrades significantly when interacting with non-cooperative patients, yet existing approaches either apply adversarial behaviors without graded severity or case-specific grounding, or reduce patient non-cooperation to a single ungraded axis, and none analyze cross-dimension interactions.

We introduce \textbf{MedDialBench}, a benchmark enabling controlled, dose-response characterization of how individual patient behavior dimensions affect LLM diagnostic robustness.
At its core is a five-dimension behavioral decomposition---Logic Consistency, Health Cognition, Expression Style, Disclosure, and Attitude---each with graded severity levels, operationalized through case-specific behavioral scripts that ensure medical plausibility and consistent patient behavior across doctor model evaluations.
This controlled factorial design enables analyses beyond prior work: while existing benchmarks support at most binary or single-axis perturbation, graded severity levels within each dimension allow dose-response profiling, and reveal interaction effects in targeted dimension combinations.

Evaluating five frontier LLMs across 7,225 dialogues (85 cases $\times$ 17 configurations $\times$ 5 models), we discover a fundamental asymmetry between two degradation pathways: \emph{information pollution} (fabricating symptoms) produces 1.7--3.4$\times$ larger accuracy drops than \emph{information deficit} (withholding information), and fabricating is the only configuration achieving statistical significance across all five models (McNemar $p < 0.05$).
Among six tested dimension combinations, fabricating is the sole driver of super-additive interaction: all three fabricating-involving pairs produce O/E ratios of 0.70--0.81 (35--44\% of eligible cases fail under the combination despite succeeding under each dimension alone), while all three non-fabricating pairs---including one involving another pollution dimension (denial)---show purely additive effects (O/E $\approx$ 1.0).
Inquiry strategy moderates deficit but not pollution: exhaustive questioning recovers withheld information, but cannot compensate for fabricated inputs.
Models exhibit qualitatively distinct vulnerability profiles, with worst-case drops ranging from 38.8 to 54.1 percentage points.
\end{abstract}

% ============================================================
% 1. INTRODUCTION
% ============================================================
\section{Introduction}

Recent studies have established that LLM diagnostic accuracy degrades substantially when models must gather information through conversation: in a randomized trial with real human participants, models achieving above 90\% accuracy on static medical exams \citep{nori2023gpt4med, saab2024gemini} dropped to below 35\% in interactive settings \citep{bean2025helpmed}.
Yet this finding raises more questions than it answers.
In real clinical encounters, patients exhibit a wide range of non-cooperative behaviors: a distrustful patient may conceal sensitive history \citep{levy2019nondisclosure}, a patient with low health literacy may hold firm misconceptions about their symptoms \citep{schillinger2021literacy}, and an anxious patient may fabricate or exaggerate complaints \citep{merckelbach2019overreport}.
These behaviors degrade the information available to the diagnosing agent through fundamentally different mechanisms---\emph{information pollution} (introducing false information) versus \emph{information deficit} (withholding true information)---yet we currently lack answers to three basic questions: \textbf{Which specific behaviors cause the most damage? How does severity affect impact? And what happens when multiple behaviors co-occur?}

Existing interactive benchmarks cannot answer these questions.
AgentClinic \citep{schmidgall2024agentclinic} supports 24 cognitive bias perturbations but applies them as ungraded binary switches, precluding dose-response analysis.
MAQuE \citep{gong2025maque} adds behavioral layers in a fixed sequential order, so that individual dimensions cannot be isolated or freely combined.
MedPI \citep{fajardo2025medpi} allows patient affect to emerge naturally, but uncontrolled emergence precludes systematic manipulation.
MedDialogRubrics \citep{gong2026meddialogrubrics} eliminates patient non-cooperation entirely to focus on diagnostic completeness.
Collectively, these approaches show \emph{that} non-cooperation degrades performance, but not \emph{which} behaviors are responsible, \emph{how} severity modulates impact, or \emph{whether} they interact.

We introduce \textbf{MedDialBench}, designed to answer precisely these questions.
At its core is a five-dimension behavioral decomposition---\emph{Logic Consistency}, \emph{Health Cognition}, \emph{Expression Style}, \emph{Disclosure}, and \emph{Attitude}---each with graded severity levels and case-specific behavioral scripts tailored to clinical details.
The controlled factorial design enables graded single-dimension sensitivity analysis, dose-response characterization, and cross-dimension interaction detection---analyses that prior designs preclude.

Our contributions: (1) A \textbf{five-dimensional behavioral framework} with graded severity, operationalized through case-specific scripts grounded in clinical communication research (\S\ref{sec:framework}).
(2) \textbf{MedDialBench}: 85 cases $\times$ 17 configurations $\times$ 5 LLMs = 7,225 dialogues with dual-judge validation ($\kappa = 0.882$).
(3) \textbf{Controlled behavioral impact analysis}: fabricating produces 1.7--3.4$\times$ larger drops than deficit and is the sole driver of super-additive interaction (35--44\% of eligible cases), while other pollution dimensions show only additive effects.
(4) \textbf{Differential vulnerability profiles}: worst-case drops range from 38.8 to 54.1 pp; inquiry strategy moderates deficit but not pollution.

% ============================================================
% 2. RELATED WORK
% ============================================================
\section{Related Work}

\subsection{Interactive Medical Dialogue Evaluation}
\label{sec:interactive}

Static medical benchmarks such as MedQA \citep{jin2021medqa}, PubMedQA \citep{jin2019pubmedqa}, and MedMCQA \citep{pal2022medmcqa} assume complete information availability.
The shift toward interactive evaluation has shown dramatic performance gaps: HELPMed \citep{bean2025helpmed} demonstrates that LLMs' 94.9\% standalone accuracy drops to 34.5\% with real human participants, and AgentClinic \citep{schmidgall2024agentclinic} finds accuracy can fall to one-tenth of static performance in multi-agent clinical simulations.
AgentClinic further introduces 24 cognitive and implicit bias perturbations, but biases are ungraded (present or absent), analyzed at the category level rather than per-bias, and no cross-bias interactions are tested.

MAQuE \citep{gong2025maque} adds behavioral layers incrementally to 3,000 simulated patients, measuring each layer's marginal effect; however, the fixed sequential order precludes arbitrary combinations and no severity gradation is provided.
MedPI \citep{fajardo2025medpi} introduces 105 evaluation dimensions with emergent patient affect, but uncontrolled emergence precludes systematic manipulation.
MedDialogRubrics \citep{gong2026meddialogrubrics} eliminates patient non-cooperation entirely to focus on diagnostic completeness.
LingxiDiagBench \citep{xu2026lingxi} benchmarks psychiatric consultation but ties patient behavior to model version rather than parameterized dimensions.
CPB-Bench \citep{cpbbench2026} annotates four behavior categories (contradiction, inaccuracy, self-diagnosis, resistance) at the utterance level, but without severity grading or factorial design.

\subsection{Patient Behavior Modeling}

AIPatient \citep{yu2024aipatient} grounds patient agents in EHR data via knowledge graphs, but personality affects only 2\% of response variation. PatientSim \citep{chen2025patientsim} defines personas through four dimensions (personality, language proficiency, recall, cognitive confusion), producing 37 combinations.
However, both characterize \emph{who the patient is} (persona attributes) rather than \emph{what the patient does} (behavioral actions)---a patient with low recall may still cooperate fully. Our framework captures adversarial \emph{behaviors} that directly manipulate information availability regardless of persona.

\subsection{Attributing Degradation to Specific Behaviors}

Prior work probes specific mechanisms---single-turn fuzzing \citep{fang2024medfuzz}, binary information completeness \citep{li2024mediq}, symptom removal \citep{wert2026q4dx}, sycophancy \citep{sycoeval2026}---but none combines graded severity with factorial design. Table~\ref{tab:comparison} summarizes key distinctions.

\begin{table}[t]
\centering
\small
\begin{tabular}{lccc}
\toprule
\textbf{Work} & \textbf{Dim.\ Isolation} & \textbf{Graded Severity} & \textbf{Interaction} \\
\midrule
AgentClinic & Partial$^\dagger$ & No & No \\
MAQuE & Partial$^\ddagger$ & No & No \\
MedPI & No & No & No \\
MedDialogRubrics & N/A & N/A & N/A \\
MediQ & No & No & No \\
Q4Dx & Yes & Partial$^\S$ & No \\
AIPatient & No & No & No \\
PatientSim & No & No & No \\
SycoEval-EM & No & No & No \\
MedFuzz & No & No & No \\
HELPMed & No & No & No \\
LingxiDiagBench & No & No & No \\
CPB-Bench & Partial$^\|$ & No & No \\
\midrule
\textbf{MedDialBench} & \textbf{Yes} & \textbf{Yes} & \textbf{Yes} \\
\bottomrule
\end{tabular}
\caption{Comparison along three analytical capabilities. $^\dagger$Binary on/off per bias. $^\ddagger$Fixed sequential addition. $^\S$Single-axis (exposure rate). $^\|$Utterance-level annotation; dialogues may contain multiple co-occurring behaviors.}
\label{tab:comparison}
\end{table}

% ============================================================
% 3. METHODS
% ============================================================
\section{Methods}

\subsection{Behavioral Framework}
\label{sec:framework}

\begin{figure}[t]
\centering
\includegraphics[width=\textwidth]{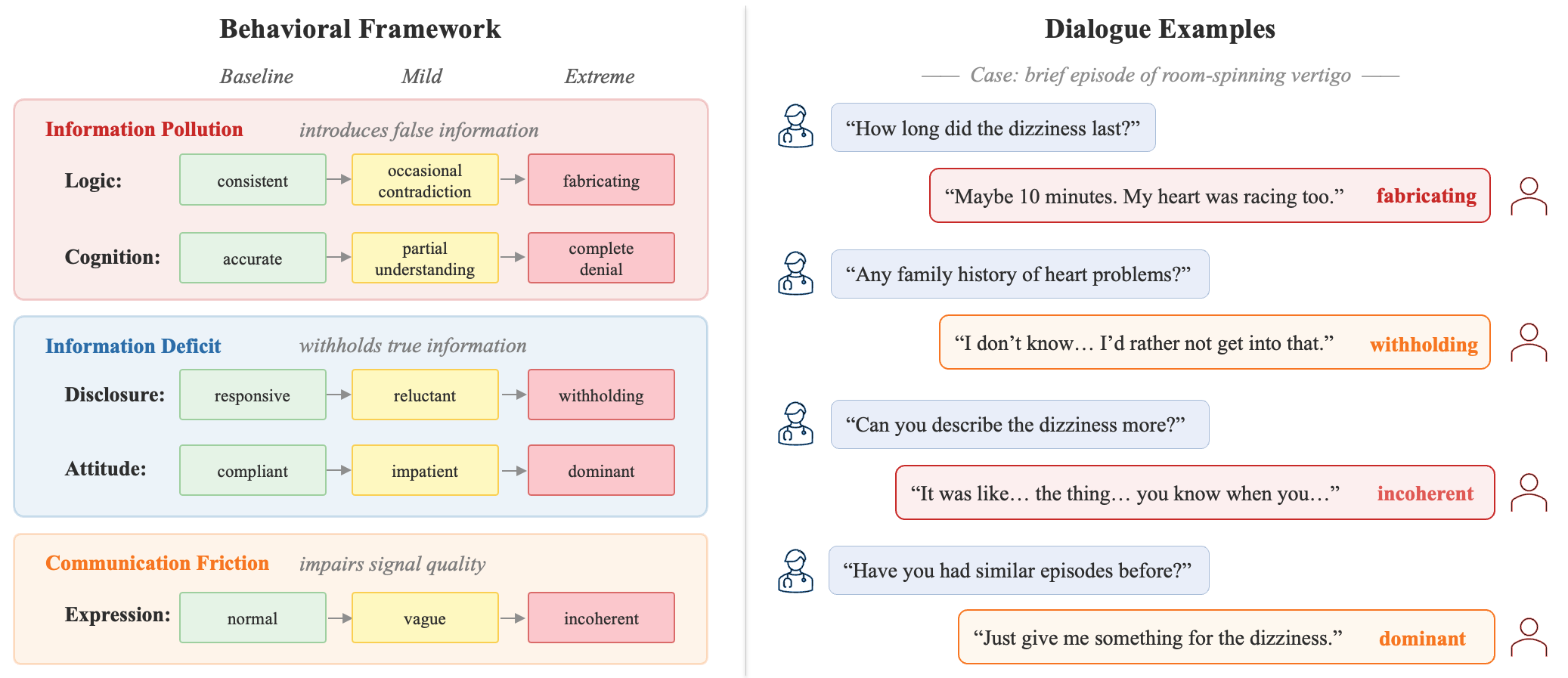}
\caption{Left: Behavioral framework with five dimensions spanning three degradation pathways, each with graded severity levels. Right: Dialogue examples illustrating adversarial behaviors from different pathways.}
\label{fig:framework}
\end{figure}

We decompose patient behavior into five controllable dimensions, each with a baseline (cooperative) level and one or more adversarial severity levels (Figure~\ref{fig:framework}), grounded in clinical communication research:
\emph{Disclosure}---60--81\% of patients report withholding medically relevant information from clinicians \citep{levy2019nondisclosure};
\emph{Logic}---symptom exaggeration and fabrication occur in 15--50\% of clinical assessments depending on context \citep{merckelbach2019overreport};
\emph{Cognition}---low health literacy leads to misattribution and misconceptions about symptoms \citep{schillinger2021literacy};
\emph{Expression}---anxiety, pain, and cognitive impairment degrade patients' ability to articulate symptoms coherently \citep{hadjistavropoulos2011pain};
\emph{Attitude}---difficult patient encounters, characterized by demanding, withdrawn, or hostile behavior, are well-documented in clinical education \citep{groves1978difficult}.

The five dimensions span three degradation pathways:
\textbf{Information pollution} (Logic, Cognition)---the patient introduces false or distorted information (fabricating symptoms, denying findings);
\textbf{Information deficit} (Disclosure, Attitude)---the patient withholds or restricts access to true information (concealing facts, deflecting questions);
\textbf{Communication friction} (Expression)---the patient's ability to convey information is impaired (off-target answers, incoherent speech).

\paragraph{Controlled factorial design.} Each experimental configuration activates at most one or two dimensions at non-baseline levels, with all others explicitly held at baseline in the patient agent's instructions. In single-dimension configurations, performance differences relative to baseline are attributable to the manipulated dimension. We note that dimensions are not strictly orthogonal---e.g., a dominant patient may incidentally reduce disclosure---but the prompt design minimizes such crosstalk by instructing the patient to cooperate fully on non-activated dimensions.

\paragraph{Case-specific behavioral scripts.} For each case and non-baseline dimension, an LLM generates a behavioral script grounded in clinical details. For example, given a case of BPPV, a fabricating script might instruct the patient to report slurred speech and visual blurring (pointing toward TIA), while a withholding script might specify concealing the brief episode duration. Scripts were manually reviewed against case details to ensure behavioral plausibility; automated behavioral adherence validation is reported in \S\ref{sec:fidelity}.

\subsection{Patient Agent}

The patient agent is an LLM (Claude Opus 4.5), selected from four previous-generation frontier models via a pilot study evaluating behavioral script adherence across all dimensions (9/9 adherence vs.\ 2/9--6/9 for alternatives; Appendix~\ref{app:patient}). We use previous-generation models for the patient agent and judge to avoid overlap with the five current-generation models under evaluation as doctors. It receives a structured prompt with four components: (1) \textbf{patient profile}---demographics and 8--16 key information items in natural language, representing all facts the patient knows; (2) \textbf{behavioral configuration}---each dimension specified as an independent instruction block with severity level and case-specific script; (3) \textbf{pacing rules}---chief complaint only in the first turn, adversarial traits emerge gradually; and (4) \textbf{full dialogue history}.

All behavioral dimensions remain \textbf{fixed} throughout each dialogue, prioritizing experimental control over ecological validity.
Disclosure serves as a ``master gate'' controlling \emph{how much} the patient reveals per turn, while other dimensions control \emph{how} information is expressed---preventing, e.g., an incoherent patient from accidentally disclosing everything in one rambling turn.

\subsection{Case Construction}

We draw from 107 OSCE cases compiled by \citet{schmidgall2024agentclinic}, each with structured clinical data (demographics, chief complaint, history of present illness, past medical history, medications, social/family history).
GPT-5.2 extracts \textbf{key information items}---discrete facts a patient could plausibly report (symptoms, onset timing, pertinent negatives, medication use)---excluding physical exam or lab findings. Each case yields 8--16 items in patient-friendly language.

We verify diagnosability empirically: each doctor model conducts one baseline consultation per case; cases where at least one model reaches the correct diagnosis (judged by Qwen3-Max) are retained, yielding \textbf{85 of 107 cases} (79.4\%). Excluded cases inherently required physical examination or imaging findings.

\subsection{Doctor Agent and Experimental Design}

The consultation proceeds in two phases: (1) \textbf{Inquiry}---the doctor asks questions until outputting \texttt{[END\_INQUIRY]} (max 20 turns), and (2) \textbf{Diagnosis}---a separate prompt asks for a specific diagnosis.

Five frontier LLMs serve as doctor agents: Gemini 3.1 Pro, GPT-5.4, Claude Opus 4.6, DeepSeek V3.2, and Qwen 3.5 Plus.\footnote{Four models used standard inference (thinking disabled). Gemini 3.1 Pro requires thinking enabled (HIGH setting), as thinking is integral to all Gemini 3.x models.}

We evaluate 17 configurations: 1 baseline + 10 single-dimension (5 dims $\times$ 2 levels) + 6 multi-dimension combinations.
The six combinations use the extreme-level dimensions with the largest single-dimension effects, as moderate levels produce near-zero effects insufficient for interaction detection: C1 (fabricating + withholding), C2 (fabricating + incoherent), C3 (fabricating + denial), C4 (withholding + dominant), C5 (withholding + incoherent), C6 (withholding + denial).
Total: 85 cases $\times$ 17 configs $\times$ 5 models = \textbf{7,225 dialogues}.

\subsection{Evaluation}

\paragraph{Semantic accuracy.} Following the LLM-as-judge paradigm \citep{zheng2023judging}, an LLM judge (Qwen3-Max, selected via a pilot comparing three candidates on 28 human-annotated dialogues; Appendix~\ref{app:judge}) determines whether the doctor's diagnosis is semantically equivalent to the ground truth ($\kappa = 0.882$ via dual-judge cross-validation with Gemini 3 Pro on 220 stratified cases).

\paragraph{Information coverage.} The fraction of key information items disclosed during dialogue.

\paragraph{Inquiry efficiency.} Coverage per turn ($\text{coverage} / \text{turns}$), measuring how effectively the doctor elicits information per unit of dialogue.

\paragraph{Misled (exploratory).} Whether an incorrect diagnosis was causally attributable to false patient information (moderate inter-judge agreement, $\kappa = 0.469$).

\paragraph{Statistical analysis.} McNemar's test for paired accuracy comparisons; bootstrap 95\% CIs (10,000 resamples). For multi-dimension combinations, we compute the \textbf{O/E ratio} (observed-to-expected ratio) under the multiplicative independence assumption \citep{bliss1939toxicity, whitcomb2023interaction}:
\begin{equation}
\text{O/E} = \frac{\text{Acc}_{\text{combo}} \times \text{Acc}_{\text{baseline}}}{\text{Acc}_A \times \text{Acc}_B}
\end{equation}
where $\text{Acc}_A$ and $\text{Acc}_B$ are the single-dimension accuracies. $\text{O/E} < 1.0$ indicates super-additive degradation (worse than independent effects predict); $\text{O/E} \approx 1.0$ indicates additive effects.

\paragraph{Patient agent behavioral adherence.}
\label{sec:fidelity}
An LLM judge (Qwen3-Max) assessed 100 stratified dialogues (all 17 configurations $\times$ 5 models) on two criteria: \emph{Activation Adherence} (activated dimensions exhibited) and \emph{Isolation Compliance} (non-activated dimensions remain at baseline).
Activation Adherence reached 94.3\% (strict full-pass; mean score 97.1\%); Isolation Compliance was 100\% across all configurations including baseline controls. Details in Appendix~\ref{app:adherence}.

% ============================================================
% 4. RESULTS
% ============================================================
\section{Results}

\subsection{Baseline Performance}

\begin{table}[t]
\centering
\small
\begin{tabular}{lcccc}
\toprule
\textbf{Model} & \textbf{Acc\%} & \textbf{Turns} & \textbf{Coverage} & \textbf{Cov/Turn} \\
\midrule
Gemini 3.1 Pro$^\dagger$ & 90.6 & 10.9 & .732 & .072 \\
GPT-5.4 & 82.4 & 17.5 & .793 & .051 \\
Claude Opus 4.6 & 77.6 & 9.4 & .744 & .083 \\
DeepSeek V3.2 & 74.1 & 18.2 & .832 & .047 \\
Qwen 3.5 Plus & 69.4 & 12.3 & .723 & .065 \\
\bottomrule
\end{tabular}
\caption{Baseline performance (85 cases). Cov/Turn = coverage per turn (inquiry efficiency). $^\dagger$Thinking enabled (HIGH).}
\label{tab:baseline}
\end{table}

Two inquiry strategies emerge: \textbf{efficient models} (Claude, Gemini; 9--11 turns) and \textbf{exhaustive models} (GPT-5.4, DeepSeek; 17--18 turns); Qwen occupies an intermediate position (12.3 turns).

\subsection{Single-Dimension Effects}

\begin{table}[t]
\centering
\small
\setlength{\tabcolsep}{4pt}
\begin{tabular}{lccccc}
\toprule
\textbf{Config} & \textbf{GPT-5.4} & \textbf{Claude} & \textbf{DeepSeek} & \textbf{Gemini} & \textbf{Qwen} \\
\midrule
Baseline & 82.4 & 77.6 & 74.1 & 90.6 & 69.4 \\
\midrule
\multicolumn{6}{l}{\emph{Expression (Friction)}} \\
\quad vague & 81.2 & 84.7 & 69.4 & 89.4 & 72.9 \\
\quad incoherent & 81.2 & 82.4 & 72.9 & 92.9 & 76.5 \\
\multicolumn{6}{l}{\emph{Cognition (Pollution)}} \\
\quad partial\_understanding & 76.5 & 75.3 & 61.2 & 85.9 & 65.9 \\
\quad complete\_denial & 71.8 & 69.4 & 55.3 & 83.5 & 62.4 \\
\multicolumn{6}{l}{\emph{Logic (Pollution)}} \\
\quad occ.\ contradiction & 78.8 & 77.6 & 65.9 & 85.9 & 70.6 \\
\quad \textbf{fabricating} & \textbf{63.5} & \textbf{55.3} & \textbf{54.1} & \textbf{60.0} & \textbf{49.4} \\
\multicolumn{6}{l}{\emph{Disclosure (Deficit)}} \\
\quad reluctant & 80.0 & 80.0 & 68.2 & 87.1 & 64.7 \\
\quad withholding & 74.1 & 68.2 & 62.4 & 76.5 & 63.5 \\
\multicolumn{6}{l}{\emph{Attitude (Deficit)}} \\
\quad impatient & 74.1 & 76.5 & 63.5 & 83.5 & 63.5 \\
\quad dominant & 75.3 & 69.4 & 61.2 & 82.4 & 64.7 \\
\bottomrule
\end{tabular}
\caption{Diagnostic accuracy (\%) by single-dimension configuration. \textbf{Bold}: fabricating, the only universally significant configuration.}
\label{tab:singledim}
\end{table}

\begin{figure}[t]
\centering
\includegraphics[width=\textwidth]{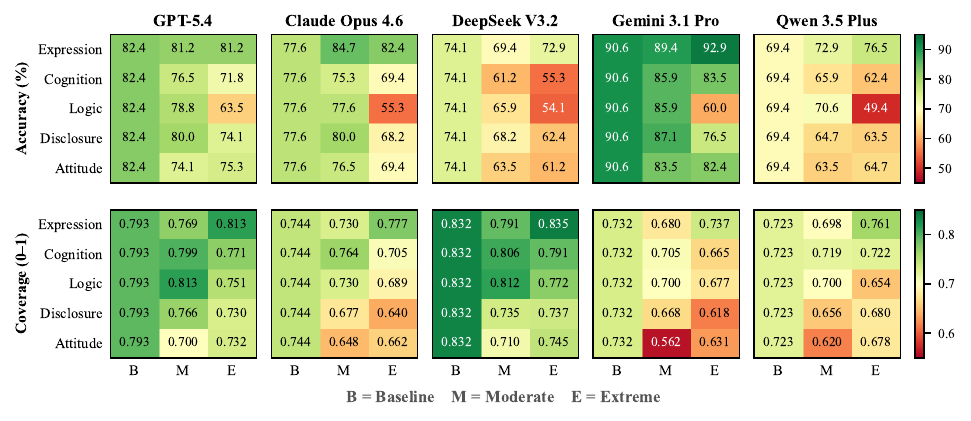}
\caption{Accuracy (top) and information coverage (bottom) heatmaps across 5 models, 5 dimensions, and 3 severity levels (B=Baseline, M=Moderate, E=Extreme). Pollution dimensions show disproportionate accuracy drops relative to coverage loss; deficit dimensions show the reverse pattern.}
\label{fig:heatmap}
\end{figure}

\begin{figure}[t]
\centering
\includegraphics[width=0.85\textwidth]{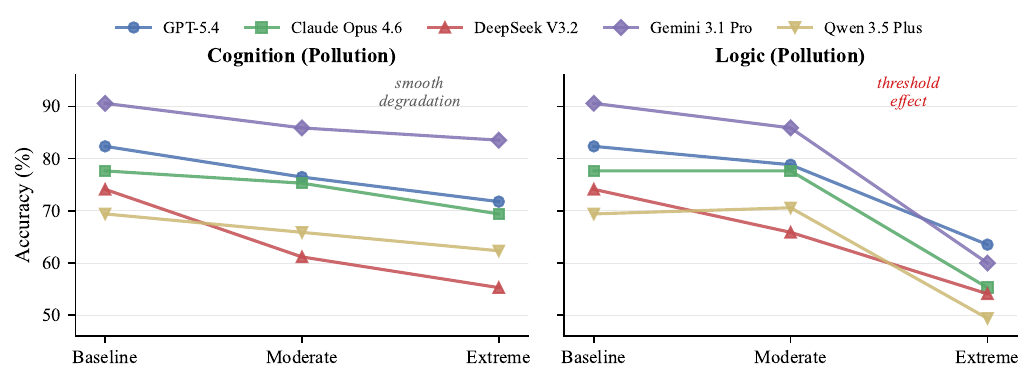}
\caption{Dose-response curves for the two pollution dimensions. Cognition (left) shows monotonic degradation with largely preserved model ranking. Logic (right) displays a threshold effect at the extreme level, with rank convergence---the best-to-worst gap narrows from 21.2 pp to 14.1 pp.}
\label{fig:dose_response}
\end{figure}

Fabricating is the only configuration achieving statistical significance (McNemar $p < 0.05$) for all five models, with accuracy drops of 18.8--30.6 pp.
Expression perturbation produces no significant degradation: incoherent patients produce verbose responses that inadvertently \emph{increase} coverage (by 0.003--0.038), and doctors compensate with fewer turns ($-$0.5 to $-$2.0 across models). Although impaired expression is a realistic clinical challenge \citep{hadjistavropoulos2011pain}, current LLMs already handle it well---communication friction alone does not degrade diagnostic accuracy when information volume is preserved. The dominant vulnerability surfaces are instead information quality (Logic: $-$22.4 pp at extreme) and information access (Disclosure: $-$9.9 pp at extreme), both of which reduce or corrupt the inputs available for diagnostic reasoning.

The multi-level design reveals dimension-specific dose-response patterns (Figure~\ref{fig:dose_response}): Cognition shows smooth monotonic degradation ($-$5.9 pp at moderate, $-$10.3 pp at extreme, 1.8$\times$ ratio), with roughly parallel curves preserving model rank order; Logic displays a threshold effect ($-$3.1 pp at moderate, $-$22.4 pp at extreme, 7.3$\times$ ratio), and strikingly, the five models \emph{converge} at extreme---the best-to-worst gap narrows from 21.2 pp at baseline to 14.1 pp under fabricating, with Gemini---the strongest baseline model---losing its lead to GPT-5.4.

The coverage heatmap (Figure~\ref{fig:heatmap}) reveals a mechanistic dissociation between pollution and deficit pathways.
Pollution dimensions produce disproportionately large accuracy drops relative to coverage loss: under fabricating, mean coverage decreases by only 0.06 yet accuracy drops by 22.4 pp.
In contrast, deficit dimensions (withholding, dominant) produce larger coverage drops ($-$0.08) but smaller accuracy drops ($-$8.2 to $-$9.9 pp).
This dissociation confirms that pollution damages through \emph{reasoning corruption}---false inputs mislead diagnostic inference---not information scarcity, explaining why more questioning cannot compensate (\S\ref{sec:discussion}).

\subsection{Multi-Dimension Interactions}

\begin{table}[t]
\centering
\small
\begin{tabular}{llccc}
\toprule
\textbf{Combo} & \textbf{Type} & \textbf{Mean Acc} & \textbf{Mean $\Delta$} & \textbf{O/E} \\
\midrule
C1: fab.+with. & Fab $\times$ Deficit & 40.2 & $-$38.6 & 0.81 \\
C2: fab.+incoh. & Fab $\times$ Friction & 40.9 & $-$37.9 & 0.70 \\
C3: fab.+den. & Fab $\times$ Pollution & 37.4 & $-$41.4 & 0.78 \\
\midrule
C4: with.+dom. & Deficit $\times$ Deficit & 67.3 & $-$11.5 & 1.09 \\
C5: with.+incoh. & Deficit $\times$ Friction & 69.2 & $-$9.6 & 0.97 \\
C6: with.+den. & Pollution $\times$ Deficit & 60.0 & $-$18.8 & 0.99 \\
\bottomrule
\end{tabular}
\caption{Multi-dimension combinations: mean accuracy across 5 models, drop from baseline, and O/E ratio ($<$1.0 = super-additive). Fab = fabricating.}
\label{tab:combos_results}
\end{table}

\begin{figure}[t]
\centering
\includegraphics[width=0.78\textwidth]{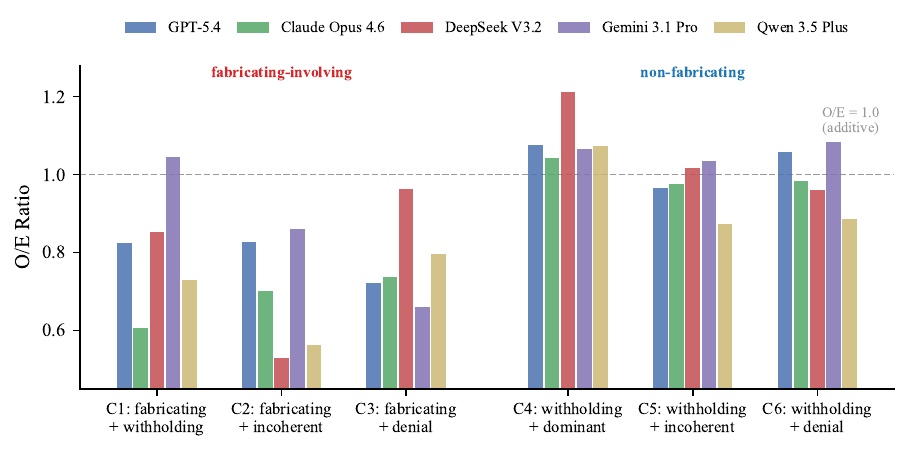}
\caption{Per-model O/E ratios for six multi-dimension combinations. Fabricating-involving combinations (C1--C3) consistently fall below 1.0 (super-additive), while all non-fabricating combinations (C4--C6) cluster at or above 1.0---including C6, which involves another pollution dimension (denial).}
\label{fig:interaction}
\end{figure}

Three key findings (Table~\ref{tab:combos_results}, Figure~\ref{fig:interaction}):
(1) \textbf{Fabricating is the sole driver of super-additivity}---all three fabricating-involving pairs (C1--C3) show O/E ratios of 0.70--0.81, while all three non-fabricating pairs (C4--C6) show O/E $\approx$ 1.0 (range 0.97--1.09). Crucially, C6 (denial + withholding) involves a pollution dimension but does \emph{not} produce super-additive degradation (O/E = 0.99), demonstrating that the effect is specific to fabricating rather than the pollution pathway in general.
(2) Among cases where the doctor succeeds under both single-dimension configurations and baseline, 40.9\% fail under fabricating-involving combinations---a failure mode invisible to single-dimension evaluation. This rate is substantially higher for fabricating pairs (\textbf{35--44\%} for C1--C3) than for non-fabricating pairs (11.1--15.6\% for C4--C6).
(3) The degree of super-additivity is model-specific: C2 (fabricating + incoherent) shows the widest spread, with DeepSeek exhibiting the strongest interaction (O/E = 0.53) and Gemini the weakest (O/E = 0.86).

\subsection{Differential Vulnerability Profiles}
\label{sec:vulnerability}

All models are more vulnerable to pollution than deficit (1.7--3.4$\times$ larger drops).
Under the most demanding deficit combination (C4: withholding + dominant), efficient models ($\leq$11 turns) suffer larger coverage drops ($-$0.164 to $-$0.179) than exhaustive models ($-$0.117 to $-$0.141).
Gemini shows a notable anomaly: under attitude-moderate (impatient), its coverage drops to 0.562 (the lowest across all model-config pairs), suggesting susceptibility to conversational pressure.
Each model has a distinct worst-case configuration: GPT-5.4 and Gemini are most vulnerable to C3 (double pollution), Claude to C1 (cross-pathway), DeepSeek and Qwen to C2 (fabricating + incoherent).
Worst-case drops range from 38.8 pp (Qwen) to 54.1 pp (Gemini).

% ============================================================
% 5. CASE ANALYSIS
% ============================================================
\section{Case-Level Analysis}
\label{sec:case}

\textbf{Super-additive interaction (Case 079, BPPV).}
Under \texttt{fabricating} alone, doctors gather enough truthful information; under \texttt{withholding} alone, questioning succeeds.
Under C1, four of five models diagnose \textbf{TIA}---withholding suppresses protective symptoms (brief duration, positional trigger) while fabricating fills the gap with a false cardiovascular narrative (slurred speech, visual blurring).

\textbf{Deficit: wrong but not misled (Case 049, Hemophilia).}
Under \texttt{withholding}, the patient conceals bruising, family history, and joint swelling; doctors diagnose \emph{post-extraction hemorrhage}---the correct interpretation of an incomplete picture, illustrating that deficit failures require better information-gathering, not improved reasoning.

\textbf{Differential robustness (Case 084, NMS).}
Under C3, the patient fabricates a classic meningitis presentation (neck stiffness, photophobia). Claude asks about medication history, discovers haloperidol, and correctly diagnoses NMS; Gemini and GPT-5.4 never ask about medications. Systematic history-taking resists diagnostic anchoring better than hypothesis-driven questioning.

% ============================================================
% 6. DISCUSSION
% ============================================================
\section{Discussion}

\label{sec:discussion}
\paragraph{Finding 1: Inquiry strategy moderates deficit but not pollution.}
Exhaustive questioning recovers withheld information (\S\ref{sec:vulnerability}), yet under pollution, misled rates are uniformly high (22--32\%) regardless of turn count.
The coverage-accuracy dissociation (Figure~\ref{fig:heatmap}) explains why: deficit reduces coverage but doctors still reason correctly from incomplete data, whereas pollution barely reduces coverage yet corrupts reasoning.
This asymmetry has a direct engineering implication: \textbf{improving questioning strategy can mitigate deficit-induced failures, but cannot address pollution-induced failures}---the latter require architectural interventions such as external verification against physical examination findings, laboratory results, or electronic health records. Case 084 (\S\ref{sec:case}) illustrates this: Claude's systematic history-taking (medication inquiry) resisted anchoring under fabrication, while hypothesis-driven models failed---suggesting that inquiry strategy classification warrants systematic study.

\paragraph{Finding 2: Fabricating is the sole driver of super-additive interaction.}
The specificity of super-additivity to fabricating (\S4.3)---absent even for denial, another pollution dimension---reveals that the mechanism is \emph{active construction of coherent false narratives}: fabricating fills information gaps with plausible but wrong clinical stories (Case 079, \S5), whereas denial merely removes information, functioning more like deficit than true pollution.

\paragraph{Implications.}
These findings define a \emph{competence envelope}: deficit-induced failures are correctable via better questioning (the information is recoverable), while pollution-induced failures require architectural interventions (external verification against physical exam, labs, or EHRs). MedDialBench delineates which conditions fall within each range.

\paragraph{Robustness $\neq$ capability.}
Gemini (90.6\% baseline) suffers the largest worst-case drop ($-$54.1 pp under C3), and the rank convergence under fabricating (\S4.2, Figure~\ref{fig:dose_response}) confirms that resistance to false inputs is largely orthogonal to baseline diagnostic ability.

% ============================================================
% 7. CONCLUSION
% ============================================================
\vspace{-1mm}
\section{Conclusion}
\vspace{-1mm}

We present MedDialBench, a benchmark for evaluating LLM diagnostic robustness under parametric adversarial patient behaviors.
Evaluation of five frontier LLMs across 7,225 dialogues reveals that fabricating is the most damaging single dimension (1.7--3.4$\times$ larger drops than deficit) and the sole driver of super-additive interaction (35--44\% of eligible cases in fabricating-involving combinations, vs.\ additive effects for all non-fabricating pairs including another pollution dimension). Each model exhibits a distinct vulnerability profile (worst-case drops: 38.8--54.1 pp).
These findings demonstrate that robustness evaluation must consider the specific adversarial landscape a diagnostic agent will face---and that benchmarks with parametric behavioral control are essential for mapping this landscape systematically.
All code, prompts, behavioral scripts, and dialogue data will be publicly released.

% ============================================================
% LIMITATIONS
% ============================================================
\vspace{-2mm}
\section*{Limitations}
\vspace{-1mm}

\paragraph{Fixed dimensions \& single patient agent.} Behavioral dimensions remain fixed per dialogue (no dynamic trajectories), and all experiments use one patient LLM (Claude Opus 4.5). This is a deliberate design choice: fixing the patient isolates doctor-side variation from confounds introduced by patient model differences. However, our findings characterize robustness \emph{as elicited by one particular patient agent}; cross-validation with alternatives would strengthen generalizability.

\paragraph{Case scale.} 85 cases is fewer than some benchmarks, but our factorial design evaluates each across 17 configurations---higher per-case cost than single-configuration designs (AgentClinic: 311 cases $\times$ 1 condition; MAQuE: 3,000 $\times$ fixed layers). Post-hoc power analysis confirms sufficient power ($1 - \beta > 0.86$) for all five models, and key findings---pollution $>$ deficit, super-additive interactions, dose-response patterns---replicate across all five independently developed LLMs.

\paragraph{Other limitations.} The benchmark excludes physical examination and laboratory testing; the open-ended diagnosis format does not capture partial credit. The Expression dimension impairs clarity but not volume; future work should test whether constraining both clarity and volume simultaneously produces degradation, as our current design cannot rule out that volume preservation explains the null result (\S4.2). The misled metric has moderate agreement ($\kappa = 0.469$). Gemini required thinking enabled; others used standard inference.

% ============================================================
% REFERENCES
% ============================================================
\bibliographystyle{plainnat}
\bibliography{references}

% ============================================================
% APPENDIX
% ============================================================
\newpage
\appendix

\section{Patient Agent Selection}
\label{app:patient}

We evaluated four frontier LLMs as patient agent candidates under a full-adversarial configuration (all behavioral dimensions at extreme levels) on two pilot cases. We defined 9 case-specific behavioral checkpoints derived from the adversarial script (e.g., ``initially conceals night sweats,'' ``eventually reveals neck lump under questioning,'' ``introduces timeline contradiction'').

\begin{table}[h]
\centering
\small
\begin{tabular}{lcc}
\toprule
\textbf{Model} & \textbf{Script Adherence} & \textbf{Content Filter} \\
\midrule
Claude Opus 4.5 & 9/9 & Pass \\
Qwen3-Max & 6/9 & Pass \\
Gemini 3 Pro & 5/9 & Pass \\
GPT-5.1 & 2/9 & Pass \\
\bottomrule
\end{tabular}
\caption{Patient agent candidate evaluation. Adherence = number of case-specific behavioral checkpoints correctly executed (out of 9). Results are consistent across both pilot cases (Case 004: DLBCL; Case 011: hemorrhoids); Case 004 shown as representative.}
\label{tab:patient_selection}
\end{table}

The selection criterion is behavioral script adherence, not general model capability. A patient agent with low adherence conflates patient execution failure with doctor robustness, undermining the causal interpretability of the factorial design.

Claude Opus 4.5 was the only model achieving full adherence. Notably, it exhibited natural information pacing under the withholding dimension: initially denying sensitive facts, then reluctantly admitting them under persistent questioning---mimicking how real patients gradually disclose under clinical pressure, even while the adversarial configuration instructs overall non-cooperation. GPT-5.1 executed only the withholding dimension, behaving as a mildly reserved but cooperative patient---unsuitable for adversarial simulation.

We additionally evaluated four smaller models (Claude Haiku 4.5, Gemini 3 Flash, Qwen 3.5 Flash, DeepSeek V3) to assess whether a cheaper alternative could achieve acceptable adherence. Under full-adversarial conditions, the best alternatives achieved 5/9 adherence (vs.\ 9/9 for Opus). In a follow-up single-dimension evaluation (the primary configuration in our experiments), the two best alternatives scored 0.56--0.59 mean adherence (averaged across 4 configurations $\times$ 2 cases, scored 0--1 per dimension) vs.\ 0.93 for Opus on the same rubric. The \texttt{incoherent} and \texttt{fabricating} dimensions were particularly degraded (0.15--0.35 vs.\ $\sim$0.9). We therefore selected \textbf{Claude Opus 4.5} for all experiments, prioritizing behavioral fidelity over cost.

\section{Judge Selection}
\label{app:judge}

After selecting the patient agent, we evaluated the remaining three models (GPT-5.1, Gemini 3 Pro, Qwen3-Max) as judge candidates. We generated 28 dialogues across 6 cases and 4 configurations, then manually annotated each with ground-truth labels: \texttt{correct} (diagnosis accuracy) and \texttt{misled} (whether false patient information causally contributed to diagnostic error).

Each judge candidate evaluated 15 selected dialogues (5 correct, 5 incorrect-not-misled, 5 incorrect-misled) with multiple repetitions. The judge prompt underwent 7 iterations, with key improvements including explicit misled-type definitions (fabrication, withholding, anchoring), passing key\_information to enable withholding detection, and adding a causal verification requirement.

\begin{table}[h]
\centering
\small
\begin{tabular}{lccc}
\toprule
\textbf{Model} & \textbf{Correct Acc.} & \textbf{Misled Acc.} & \textbf{Reps} \\
\midrule
\textbf{Qwen3-Max} & \textbf{100\%} & \textbf{100\%} & 5 \\
Gemini 3 Pro & 100\% & 80\% & 1 \\
GPT-5.1 & 100\% & 20\% & 3 \\
\bottomrule
\end{tabular}
\caption{Judge candidate evaluation on 15 human-annotated dialogues. Correct Acc.\ = agreement with human labels on diagnostic accuracy. Misled Acc.\ = agreement on whether fabricated/withheld information causally contributed to diagnostic error.}
\label{tab:judge_selection}
\end{table}

All three candidates achieved perfect diagnostic accuracy judgment, but differed sharply on misled discrimination. GPT-5.1 detected only withholding-type errors (Type B) and failed on fabrication/anchoring (Types A/C). Gemini 3 Pro produced 2 false positives and 1 false negative on boundary cases. Qwen3-Max correctly identified all three misled types across 5 repeated evaluations, including boundary cases where the doctor partially adopted the patient's false narrative.

We selected \textbf{Qwen3-Max} as the primary judge. Gemini 3 Pro served as the secondary judge for cross-validation ($\kappa = 0.882$ on semantic accuracy across 220 stratified cases). Since agreement was high, all experiments use the primary judge's labels without manual adjudication. For the misled metric, inter-judge agreement was substantially lower ($\kappa = 0.469$), reflecting the inherent subjectivity of causal attribution; we therefore designate misled as an exploratory metric in the main text.

\section{Behavioral Adherence Details}
\label{app:adherence}

Having selected Claude Opus 4.5 as the patient agent (Appendix~\ref{app:patient}), we further validated its behavioral adherence across the full experimental set. We evaluated 100 dialogues stratified across all 17 configurations and 5 doctor models. An independent LLM judge (Qwen3-Max) assessed each dialogue on two criteria:

\paragraph{Activation Adherence.} Whether each activated (non-baseline) dimension was exhibited in the patient's responses. For dual-dimension configurations, each dimension was scored independently.

\paragraph{Isolation Compliance.} Whether non-activated dimensions remained at baseline behavior (no behavioral artifacts).

\begin{table}[h]
\centering
\small
\begin{tabular}{lcccc}
\toprule
\textbf{Config Type} & \textbf{N} & \textbf{Activation} & \textbf{Isolation} & \textbf{Full Pass} \\
\midrule
Baseline & 30 & N/A & 100\% & 100\% \\
Single-dimension & 44 & 95.5\% & 100\% & 95.5\% \\
Dual-dimension & 26 & 92.3\% & 100\% & 92.3\% \\
\midrule
All non-baseline & 70 & 94.3\% & 100\% & 94.3\% \\
\bottomrule
\end{tabular}
\caption{Behavioral adherence results. Activation = strict full-pass rate (all activated dimensions exhibited). Mean activation score (with partial credit) was 97.1\%.}
\label{tab:adherence_detail}
\end{table}

Among the four non-full-pass cases: two were single-dimension failures (activated dimension not exhibited) and two were partial executions in dual-dimension configurations (one dimension activated, the other not). Isolation Compliance was 100\% across all configurations, confirming that the patient agent introduces no behavioral artifacts beyond what is explicitly instructed.

\section{Case-Specific Behavioral Script Examples}
\label{app:scripts}

Each case-specific script provides concrete behavioral instructions grounded in clinical details. We illustrate three examples spanning different dimensions and fabrication modes.

\paragraph{Logic = fabricating (Case 079).}
The script instructs the patient to introduce plausible contradictions:
(1) Initially denies vomiting (``I didn't actually throw up''), later admits to vomiting several times;
(2) First estimates the episode lasted ``10--15 minutes,'' later agrees it was closer to 3 minutes;
(3) Early on says ``I still feel a little off,'' later states ``I feel back to normal.''
These contradictions are designed to be individually resolvable through persistent questioning, but collectively they introduce noise that may redirect the doctor toward conditions with longer, more variable episodes (e.g., vestibular migraine, TIA).

\paragraph{Disclosure = withholding (Case 079).}
The script specifies 9 of 15 key information items as initially hidden. The patient minimizes nausea and vomiting (embarrassment), withholds the positional trigger and brief duration (considers them irrelevant), and initially answers vaguely about headache and hearing loss (``I'm not sure''). Items are revealed only after the doctor asks directly, explains clinical relevance, and reassures the patient.

\paragraph{Fabrication via invented symptoms (Case 084, NMS, under C3).}
While the Case 079 fabricating script above operates through denial and exaggeration of existing facts, fabrication can also manifest as inventing entirely new symptoms. Under C3 (denial + fabricating), the Case 084 patient---whose true presentation is fever, rigidity, and confusion from neuroleptic malignant syndrome---spontaneously fabricated: ``I've had this pounding headache, like an 8 out of 10, and my neck kind of hurts when I look down'' and ``I've been running to the bathroom a lot, like watery stuff.'' None of these symptoms appear in the case's key information. The fabricated headache + neck stiffness + fever construct a classic meningitis presentation, leading four of five doctors to diagnose meningitis rather than NMS (\S5).

\section{Case 079: Super-Additive Interaction Detailed}
\label{app:case079}

Case 079 (BPPV) illustrates the super-additive interaction mechanism discussed in \S5. Under each single-dimension configuration, all five models diagnose correctly; under C1 (withholding + fabricating), four of five fail.

\begin{table}[h]
\centering
\small
\begin{tabular}{lcccl}
\toprule
\textbf{Config} & \textbf{Correct?} & \textbf{Models correct} & \textbf{Models wrong} & \textbf{Wrong diagnosis} \\
\midrule
Baseline & 5/5 & All & --- & --- \\
Fabricating & 5/5 & All & --- & --- \\
Withholding & 5/5 & All & --- & --- \\
C1 (both) & 1/5 & Qwen & 4 & TIA \\
\bottomrule
\end{tabular}
\caption{Case 079 diagnostic outcomes across configurations.}
\label{tab:case079}
\end{table}

\paragraph{Mechanism.} Under fabricating alone, the patient introduces contradictions (denies then admits vomiting, inflates episode duration) but the doctor still elicits the critical BPPV features: brief duration, positional trigger, full resolution. Under withholding alone, the patient initially conceals these features but reveals them under persistent questioning.

Under C1, the two dimensions create a synergistic trap: withholding suppresses the protective truthful information (brief 3-minute duration, positional trigger at bedtime, complete resolution, no neurological deficits), while fabricating fills the resulting gap with a coherent but false cardiovascular narrative. The fabricating script's inflated duration (``10--15 minutes'') is no longer corrected because the withholding script prevents the patient from volunteering the true 3-minute duration. Four models converge on \textbf{TIA}---a clinically plausible diagnosis given the fabricated presentation (sudden onset, longer duration, vague neurological concerns) in a 59-year-old male.

Qwen 3.5 Plus is the sole model to diagnose correctly. Its key move was asking specifically: ``did this episode start immediately after you changed your head position, such as rolling over in bed?''---which elicited the positional trigger despite the patient's fabricated cardiovascular narrative. Qwen then performed a positional maneuver (Dix-Hallpike) that reproduced the vertigo, confirming BPPV.

\section{Complete Results}
\label{app:results}

Table~\ref{tab:full_results} presents diagnostic accuracy for all 17 configurations across all 5 models, with McNemar's test significance markers.

\begin{table}[h]
\centering
\scriptsize
\setlength{\tabcolsep}{2.5pt}
\begin{tabular}{l|c|cc|cc|cc|cc|cc|cccccc}
\toprule
& & \multicolumn{2}{c|}{\emph{Expr.}} & \multicolumn{2}{c|}{\emph{Cog.}} & \multicolumn{2}{c|}{\emph{Logic}} & \multicolumn{2}{c|}{\emph{Disc.}} & \multicolumn{2}{c|}{\emph{Att.}} & \multicolumn{6}{c}{\emph{Combos}} \\
\textbf{Model} & \textbf{BL} & \textbf{vag} & \textbf{inc} & \textbf{par} & \textbf{den} & \textbf{occ} & \textbf{fab} & \textbf{rel} & \textbf{wit} & \textbf{imp} & \textbf{dom} & \textbf{C1} & \textbf{C2} & \textbf{C3} & \textbf{C4} & \textbf{C5} & \textbf{C6} \\
\midrule
GPT-5.4 & 82.4 & 81.2 & 81.2 & 76.5 & 71.8$^*$ & 78.8 & 63.5$^*$ & 80.0 & 74.1 & 74.1 & 75.3 & 47.1$^*$ & 51.8$^*$ & 40.0$^*$ & 72.9 & 70.6 & 68.2 \\
Claude & 77.6 & 84.7 & 82.4 & 75.3 & 69.4 & 77.6 & 55.3$^*$ & 80.0 & 68.2 & 76.5 & 69.4 & 29.4$^*$ & 41.2$^*$ & 36.5$^*$ & 63.5$^*$ & 70.6 & 60.0$^*$ \\
DeepSeek & 74.1 & 69.4 & 72.9 & 61.2$^*$ & 55.3$^*$ & 65.9 & 54.1$^*$ & 68.2 & 62.4$^*$ & 63.5$^*$ & 61.2$^*$ & 38.8$^*$ & 28.2$^*$ & 38.8$^*$ & 62.4$^*$ & 62.4 & 44.7$^*$ \\
Gemini & 90.6 & 89.4 & 92.9 & 85.9 & 83.5 & 85.9 & 60.0$^*$ & 87.1 & 76.5$^*$ & 83.5 & 82.4 & 52.9$^*$ & 52.9$^*$ & 36.5$^*$ & 74.1$^*$ & 81.2 & 76.5$^*$ \\
Qwen & 69.4 & 72.9 & 76.5 & 65.9 & 62.4 & 70.6 & 49.4$^*$ & 64.7 & 63.5 & 63.5 & 64.7 & 32.9$^*$ & 30.6$^*$ & 35.3$^*$ & 63.5 & 61.2 & 50.6$^*$ \\
\bottomrule
\end{tabular}
\caption{Diagnostic accuracy (\%) across all 17 configurations and 5 models. BL = baseline. Single-dim abbreviations: vag = vague, inc = incoherent, par = partial understanding, den = complete denial, occ = occasional contradiction, fab = fabricating, rel = reluctant, wit = withholding, imp = impatient, dom = dominant. $^*$McNemar $p < 0.05$ vs.\ baseline.}
\label{tab:full_results}
\end{table}

% ============================================================
% CHECKLIST (included for NeurIPS submission; commented out for arXiv)
% ============================================================
% \input{checklist}

\end{document}